\documentclass{ecai}
\usepackage{graphicx}
\usepackage{latexsym}
\usepackage{amsmath,amsfonts,amssymb}
\usepackage{multirow}
\usepackage{xcolor}
\usepackage{subcaption}
\usepackage{booktabs}
\usepackage{hyperref}

\begin{document}

\begin{frontmatter}

\title{Specializing Small Language Models towards Complex Style Transfer via Latent Attribute Pre-Training}

\author[1]{\fnms{Ruiqi}~\snm{Xu}\thanks{Equal Contribution.}\orcid{0009-0001-7874-0725}}
\author[1]{{\fnms{Yongfeng}~\snm{Huang}}\thanks{Corresponding Author. Email: dengdeng9209@gmail.com}\protect\footnotemark[1]}
\author[1]{{\fnms{Xin}~\snm{Chen}}\thanks{Corresponding Author. Email: 1774885528a@gmail.com.}}
\author[2]{{\fnms{Lin}~\snm{Zhang}}}

\address[1]{Platinum AI Inc., China}
\address[2]{Symbiotic Matrix, China}

\begin{abstract}
In this work, we introduce the concept of complex text style transfer tasks, and constructed complex text datasets based on two widely applicable scenarios. Our dataset is the first large-scale data set of its kind, with 700 rephrased sentences and 1,000 sentences from the game Genshin Impact. While large language models (LLM) have shown promise in complex text style transfer, they have drawbacks such as data privacy concerns, network instability, and high deployment costs. To address these issues, we explore the effectiveness of small models (less than T5-3B) with implicit style pre-training through contrastive learning. We also propose a method for automated evaluation of text generation quality based on alignment with human evaluations using ChatGPT. Finally, we compare our approach with existing methods and show that our model achieves state-of-art performances of few-shot text style transfer models.
\end{abstract}

\end{frontmatter}

\section{Introduction}
Text style transfer is a task in natural language generation that involves modifying the style of a given text while preserving its content. It has a wide range of applications, including conversational assistant with customized persona \cite{li2016personabased}, writing assistant \cite{syed2020adapting}, automatic text simplification \cite{jin2020hooks}, text debiasing \cite{pryzant2019automatically}, and censoring offensive language \cite{nogueira-dos-santos-etal-2018-fighting}. However, traditional approaches to text style transfer often rely on parallel corpora, which may be unavailable or require significant manual effort to collect and annotate \cite{jhamtani-etal-2017-shakespearizing}. Recent work has shown promising results in unsupervised methods that offer an alternative solution by leveraging large amounts of unpaired text data without the need for explicit parallel annotation \cite{lample2018multipleattribute, li-etal-2018-delete}. However,  unsupervised methods often suffer from the lack of explicit control over the generated text's output and may produce outputs that do not adhere to the desired style, making them less suitable for specific style transfer tasks. Previous research has also been concentrated on transferring text across simple styles like sentiment and politeness, while there have been few studies on more complex text style transfers like personality, creativity, and conciseness. 

In this work, we first define the complex text styles as styles that are hardly distinguishable from each other except by professionals working in the relevant fields. For example, the lines from two similar characters in a video game are complex text styles as only the designers of the characters may discern the subtle differences between the personalities of the two figures. The high standard required for labeling texts of complex styles makes crowd-sourcing infeasible to generate high-quality datasets of complex text styles even in non-parallel settings. To tackle this problem and facilitate the study of text style transfer models, we picked two complex styles of interest, authorship and creativity, and constructed two large-scale datasets for benchmarking complex style transfer power of language models. 

While large models like LLM have shown promise in complex text style transfer, they have drawbacks such as data privacy concerns, network instability, and high deployment costs. To address these issues, we explore the effectiveness of small models (less than T5-3B) with implicit style pre-training through contrastive learning. With introducing the concept of specialization, we develop a high-efficiency text style generator that can be deployed offline with low costs.

Automatic evaluation of generation results is a challenging task in complex text style transfer because it requires an objective and reliable way to measure the quality of generated text. To address this challenge, we proposed a novel evaluation method based on ChatGPT, which involves generating a response from ChatGPT given a prompt that asks ChatGPT to classify the generated text, and then comparing the response with human evaluations of the same texts.

To validate the effectiveness of our evaluation method, we conducted experiments on both simple and complex datasets, and found that the alignment between ChatGPT and human evaluation reached 98$\%$ and 93$\%$ respectively. This result indicates that our proposed method is reliable and can provide a useful tool for automated evaluation of complex text style transfer models. We also measured the accuracy of our evaluation method using traditional metrics (SacreBLEU), and discuss the alignment of the accuracy metrics in the experiments on both simple and complex datasets.

The contributions of this paper can be summarized as follows: (1) We introduced the concept of complex text style transfer, and constructed two benchmark datasets for evaluating complex text style transfer models; (2) We proposed an implicit style pre-training method for small-scale models, which achieved state-of-art performances of few-shot approaches on complex text style transfer tasks and reached comparable performances to large language models; and (3) We introduced an automatic evaluation method for complex text style transfer based on ChatGPT, which provides a more objective and efficient way than previous metrics to evaluate complex text style transfer models based on human evaluation experiment results. Our work can be accessed at the following repository: \href{https://github.com/ruiqixu37/BTTS_ECAI2023}{code}.

\section{Methods}
\subsection{Preliminaries}
\begin{figure*}[t]
  \centering
  \begin{subfigure}[b]{0.59\linewidth}
    \includegraphics[width=\linewidth]{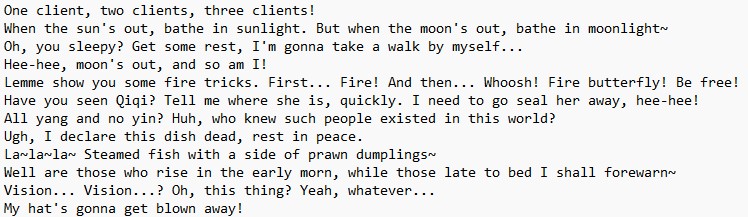}
    \label{fig:hutao_sample}
  \end{subfigure}
  \hfill
  \begin{subfigure}[b]{0.39\linewidth}
    \includegraphics[width=\linewidth]{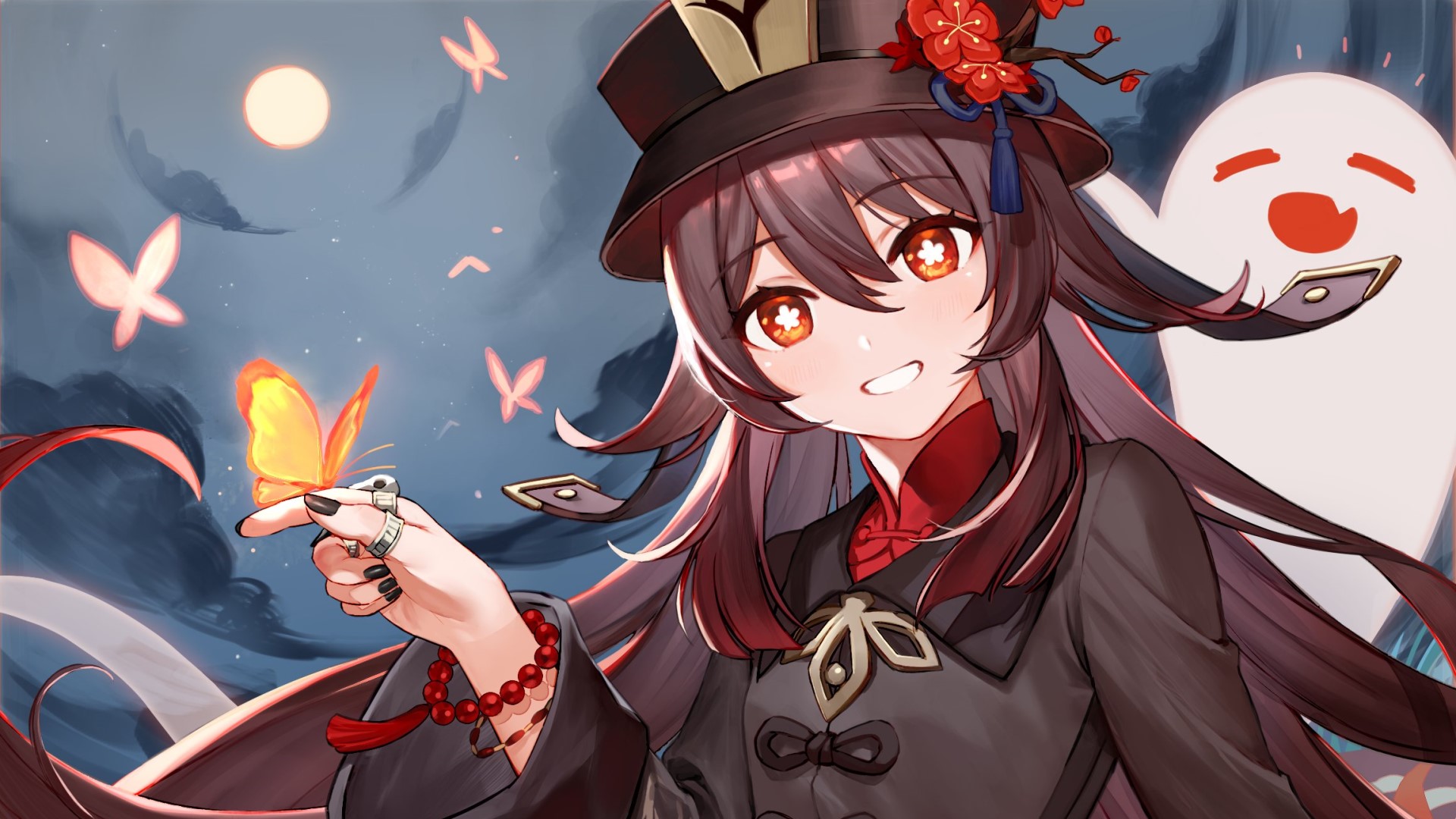}
    \label{fig:hutao_image}
  \end{subfigure}
  
  \begin{subfigure}[b]{\linewidth}
    \includegraphics[width=\linewidth]{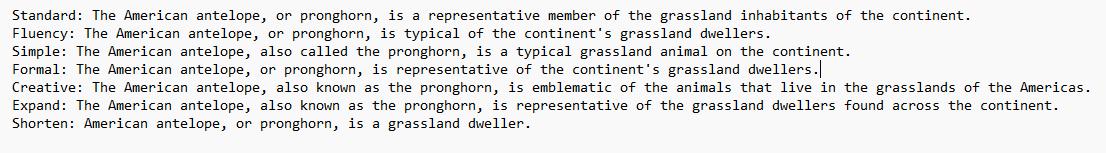}
    \label{fig:rephrase_sample}
  \end{subfigure}=
  \begin{tabular}{|c|c|c|c|}
    \hline
    Dataset & Size & Parallel & Characters/Styles\\
    \hline
    Genshin & 3,120 & No & 48\\
    Rephrase & 1,400 & Yes & 7\\
    \hline
  \end{tabular}
  \caption{Examples and specifications from our new complex style Genshin and Rephrase dataset. \textit{\textbf{Top:}} Selected lines from the Genshin dataset by \textit{Hu Tao}, a female character in the game \textit{Genshin Impact} that is commonly perceived to be energetic and outgoing. \textit{\textbf{Middle:}} Example sentence describing the American antelope in seven different styles from the Rephrase dataset. \textit{\textbf{Bottom:}} Specifications for the two datasets. It should be noted that some lines are shared by all the characters in the Genshin dataset, so it is possible to make a parallel-corpus version of our Genshin dataset. In our study, we leverage the parallel datasets in the same way as non-parallel datasets.}
  \label{datasetfiguretable}
\end{figure*}

\subsubsection{Problem Definition}
Text style transfer is the task of automatically transforming the style of a given text into a different target style while preserving its content and meaning. Some text styles are well-defined and characterized by specific attributes that are easily recognizable or distinguishable from other styles, which we term as simple text styles. For example, the happy text style is characterized by a cheerful tone, a lively sentence structure, and the use of positive words and expressions such as "happy", "joyful", "exciting", and "amazing". On the other hand, the sad text style is characterized by the use of negative words and expressions such as "gloomy", "depressed", "heartbroken", "lonely", etc.. These text styles are relatively straightforward and commonly recognized by most people. As a result, researchers can easily obtain labeled datasets for simple text style transfers by using crowdsourcing methods. 

While simple text style are common in the real world and are easily recognizable, more complex styles can pose a greater challenge and are often of greater application value. For example, in the legal domain, converting verbose legal documents into plain language versions could increase accessibility for non-experts, while in the medical field, transferring complex medical jargon into layman's terms could improve patient understanding and compliance.

In this work, we give a preliminary definition to the complex text style, which refers to the style of a given text that is difficult for non-experts to discern and categorize. Such complex styles may include personality, domain-specific jargon, or other highly specialized terminology. The nature of these styles makes it difficult to rely on crowdsourcing to label the texts, as only experts in the relevant field may be able to accurately distinguish between different styles. Therefore, the challenge of complex text style transfer lies in developing effective models that can capture and transfer these complex stylistic nuances without relying on extensive labeled data. 

\subsubsection{Dataset Descriptions}
To study the complex style transfer tasks, we construct two large-scale datasets based on two domains, personality and creativity, for bench marking the complex style transfer power of language models. The descriptions for each dataset are as below, and the samples and specifications are shown in Table \ref{datasetfiguretable}. 

\textbf{Genshin} is a collection of dialogues spoken by characters in the video game Genshin Impact. The dataset includes lines spoken by over 48 characters, with each character having distinct personalities and speaking styles. Each character has 50-80 lines of non-parallel dialogue, which means certain characters have unique lines that do not have corresponding lines from other characters.

\textbf{Rephrase} consists of parallel corpus of 200 English sentences in seven different styles (Standard, Fluency, Formal, Simple, Creative, Expand, Shorten). To produce the dataset, we first collect 200 sentences from the Internet with uncorrelated content. Then we paraphrase the sentences with QuillBot, a powerful online paraphrasing tool. For each style paraphrasing process, QuillBot is instructed to prioritize preserving semantic content over style transformation.

To validate the effectiveness of our model and make the experiments directly comparable to previous approaches, we also consider two simple groups of styles. The first is the Amazon sentiment dataset \cite{ni-etal-2019-justifying}, which consists of reviews on Amazon that are labeled either positive or negative. The second dataset we use is the Grammarly’s Yahoo Answers Formality Corpus (GYAFC) \cite{gyafcrao2018} dataset, containing a total of 110K informal / formal sentence pairs. 

\subsection{Latent Style Space Pre-Training}
\begin{figure*}[t!]
    \centering
    \includegraphics[width=0.7\textwidth]{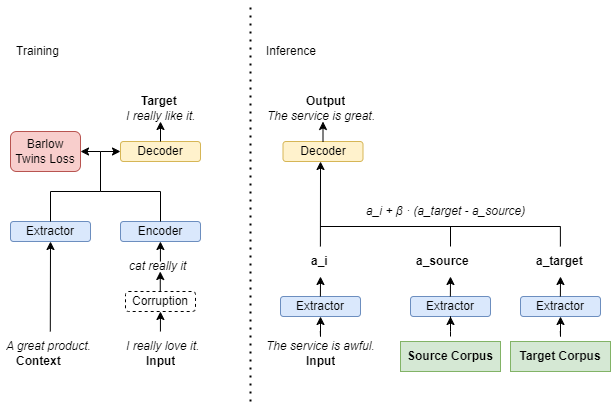}
    \caption{Model Architecture of BTTS. It consists of three parts: the Encoder, Decoder, and Extractor, which are built using transformer stacks initialized from pretrained T5. The model is trained to reconstruct a corrupted input while being conditioned on a fixed-width "style vector" extracted from the previous sentence. During inference, a new style vector is generated using "targeted restyling" by adding a directional beta to the extracted style of the input text. The decoder is also given additional conditioning through stochastic tuning ranges, which provides fine-grained control during inference. }
    \label{fig:arch-model}
\end{figure*}

Fig. \ref{fig:arch-model} describes our overall model architecture. Our work is closely reminiscent of \cite{riley2021textsettr}, which uses a large pretrained language model to learn style representations. Our work differ in the way that we include a contrastive loss that measures the similarity between the input embeddings and extractor embeddings, which allows the style extractor to capture more precise text style representations\cite{chen2020contrastiveloss, Wang_2021_CVPR}. We describe the model architecture in details below.

Our model is designed with two simple observations: (1) the pretrained large language models likely already contain powerful style representation; (2) style tends to remain unchanged within a given piece of corpus. Our model includes a encoder-decoder module based on Text-to-Text Transformer (T5) \cite{https://doi.org/10.48550/arxiv.1910.10683}. During training, we use corrupted versions of inputs and instructs the model to restore the original sentences, thereby resulting in a reconstruction task. We expect the corruption strategies to eliminate the style from original sentences and hence force our model to discover the style representation during reconstruction. Such reconstruction task also aligns with how the original T5 model is trained. 

Inspired by \cite{riley2021textsettr}, our model consists of an additional style extractor module. During our experiments, the architecture of the extractor is the same as the one used in the encoder-decoder module (either T5-base or T5-large), and its input is an uncorrupted sentence preceding the target. The output of the extractor is added to that of the encoder, which is then fed into the decoder to produce the final output. The weights of our model are initialized with those of a pretrained model, but the weights are not fixed during training.

To improve the learning power of our model, we introduce Barlow Twins loss 
\cite{https://doi.org/10.48550/arxiv.2103.03230} that measures the 
similarity of learned representations of context and target sentences. The 
Barlow Twins loss measures the difference between the empirical cross-
correlation matrix $\mathcal{C} = z_{A}^{\intercal} z_{B} \in 
\mathbb{R}^{D \times D}$ of the embeddings.
Given two sets of inputs ${x}_A$ and ${x}_B \in \mathbb{R}^{N\times C \times H \times W}$ within the same data class and matrices of their corresponding embedding vectors ${z}_A$ and ${z}_B \in \mathbb{R}^{N \times D}$ given by the encoder, the loss is defined as:

\begin{equation}
    \textsc{BT}({z}_A, {z}_B) =
    \sum_{0 \leq i < D}(1 - \mathcal{C}_{ii})^2 +
    \delta 
    \sum_{0 \leq i < D} \sum_{0 \leq j < D, \, j\neq i} \mathcal{C}_{ij}^2   
\label{barlowtwinsloss}
\end{equation}

where $D$ is the dimension of the embedding space, and $\mathcal{C}_{ij}$ stands for the elements in the empirical cross-correlation matrix between ${z}_A$ and ${z}_B$. $\delta$ is a non negative hyper-parameter, which we find the optimal value to be around $1 * 10^{-4}$.

The Barlow Twins loss is able to evaluate the similarity between the embedding vectors, and encourages different features of the embedding vectors to be less correlated. In this paper, we apply Barlow Twins on the embeddings of the extractor modules to enforce the extractor to learn an expressive latent space that captures the implicit attributes within complex styles. We apply Barlow Twins on two levels. For the sentence-level, We use the sentence directly preceding the input sentence as the context sentence, run the extractor module on the context sentence and the input sentence, and minimize the Barlow Twins loss between the two. For the paragraph-level, we use all the sentences in the corpus except for the input sentence itself as context sentences, then calculate the average of the Barlow Twins loss for each pair of the input-context pairs.
\subsubsection{Training Objective}
The final training objective is the combination of the reconstruction loss and the Barlow Twins loss. We use cross entropy between the decoder output and the target sentence as the reconstruction loss. Formally, define the corruption strategies as $f$, extractor as $ext$, encoder as $enc$, and decoder as $dec$. For a pair of sentence $s_{target}$ and its preceding sentence $s_{context}$ (used as Context), the reconstructed sentence is given by
$$s_{\text{recon}} = \text{dec}(\text{enc}(f(s_{\text{target}})) + \text{ext}(s_{\text{context}}))$$
The training objective is given by
\begin{equation}
    L= \text{CE}(s_{\text{recon}}, s_{\text{target}}) 
    + \lambda \cdot \text{BT}(\text{ext}(s_{\text{context}}), \text{ext}(s_{\text{recon}}))
\label{trainobj}
\end{equation}
where $\lambda$ is a hyperparameter.

\subsubsection{Inference}
Our model applies a few-shot approach during inference time. To transfer a sentence $i$ from source style attribute $a_{src}$ to target style attribute $a_{\text{target}}$, we assume that a small number of sentences are available for each style. We derive the style representation by running the extractor on each style set of sentences and averaging the extractor outputs, giving style attributes $a_{\text{src}}$ and $a_{\text{target}}$. We also infer the basis style attribute vector $a_{i}$ by running the extractor on the sentence $i$ itself and perform a linear transformation in the style vector space, producing the style difference vector $a_{diff}$ as 
$$a_{\text{diff}} = a_{i} + \beta \cdot (a_{\text{target}} - a_{\text{src}})$$

where $\beta$ is a hyperparameter. The style difference vector is then added back to the encoder-decoder module, resembling the architecture during training phase and producing the final transferred sentence.

In practice, we find too large $\beta$ forces the transferred sentence to lose semantic meaning of the original sentence, while small $\beta$ deactivates the style transfer. We find the optimal $\theta$ to be around $[1, 20]$, depending on the specific attributes of interests. 

\subsection{Automatic Evaluation Procedure for Complex Style Transfer Tasks}
\label{evalmetric}
\textbf{Limitations of Previous Approaches} Automatic evaluation of text style transfer tasks usually involve measuring the transferred style strength and semantic preservation of the outputs. The evaluation of complex text style tasks requires experts and cannot be achieved through crowd-sourcing, making automated evaluation a challenge. Previous research efforts in this area have relied on measuring the transferred style strength with a separately trained style classifier based on BERT \cite{DBLP:journals/corr/abs-1810-04805} and or its variants such as RoBERTa-Large \cite{DBLP:journals/corr/abs-1907-11692} and DeBERTa-v3-Large \cite{he2021debertav3, he2021deberta}.

While this approach has shown promise in some contexts, it has been observed that complex text style transfer often involves multiple categories and low-data scenarios, resulting in a dearth of available samples. This would create difficulties for the training of BERT classifiers, which requires enormous labeled style datasets. As a result, the accuracy of BERT-based automated evaluation methods in complex style transfer tasks is low and can be comparable to random results, as shown in Table \ref{accuracytable}.

\begin{table}
\centering
\caption{Classification Accuracy on simple and complex datasets by BERT-based models and OpenAI's GPT models. The simple accuracy is reported as the average of a model's classification accuracy on the formality and sentiment dataset, while the complex accuracy are the average of a model's classification accuracy on the Genshin and Rephrase dataset.}
\begin{tabular}{ c c c c } 
 \hline
  & \textbf{Name} & \textbf{Simple} & \textbf{Complex} \\ 
 \hline
 \multirow{3}{2.5em}{\textbf{Bert}}
 & BERT & 86.7 & 20.2 \\
 & RoBERTa & 89.7 & 26.7 \\
 & DeBERTa & 92.6 & 31.2 \\
 \hline
 \multirow{3}{2.5em}{\textbf{LLM}}
 & gpt-3.5-turbo & 99.3 & 75.6 \\
 & text-davinci-003 & 99.5 & 73.3 \\
 & code-davinci-002 & 94.4 & 65.4 \\
 \hline
\end{tabular}
\label{accuracytable}
\end{table}

\textbf{ChatGPT-based Accuracy Evaluation} Recent research on large language models (LLMs) such as ChatGPT has highlighted their potential for breakthroughs in NLP understanding and generation capabilities, even under conditions of few-shot or zero-shot scenarios. Building on these developments, we propose a set of LLM-based automatic evaluations for complex text style tasks, as illustrated in Table \ref{accuracytable}. The proposed methodology involves constructing dataset-dependent prompts to guide the evaluation process.

To test the effectiveness of this approach, we conducted experiments on both complex and simple datasets. The results, as presented in Table \ref{accuracytable}, demonstrate that the automatic evaluation capabilities of the LLM-based models are significantly stronger than those of simpler language models (SLMs). The high accuracy achieved by the LLM-based models attests to the efficacy of this approach for automated evaluation of complex text style tasks. 

Given the strength of ChatGPT among LLMs, it was selected as the preferred model for evaluating accuracy in our study. We used the gpt-3.5-turbo model as the default classifier because it is one of the most powerful and affordable language model with publicly available APIs as of date. The current price for gpt-3.5-turbo is $\$$0.002 per one thousand tokens and the whole evaluation process costs less than ten dollars. We also compared the classification accuracy of gpt-3.5-turbo with two other OpenAI models, text-davinci-003 and code-davinci-002 and the results are reported in Table \ref{accuracytable}. The results show that all three models achieve nearly perfect accuracy on the simple datasets, while more than doubling the performance of DeBERTa-v3-large, the best performing Bert model, on complex datasets. Overall, the findings of this study highlight the potential of LLM-based automated evaluations for complex text style tasks, especially in contexts where sample sizes are limited. We encourage future research in this area to explore the applicability of these approaches in various NLP domains and investigate the potential of other LLMs for automated evaluation tasks.

\textbf{Other Metrics} Finally, we utilize SacreBLEU \cite{post-2018-call} to measure the content preservation between the output and the input, and also report the geometric mean ("G-score") of accuracy and SacreBLEU as an overall evaluation of the model performance following \cite{xu-etal-2018-unpaired}.

\section{Experiments}
In this section, we examine BTTS on several datasets. We present the main quantitative results in Sec \ref{quantresult}, analysis on the effects of contrastive loss in Sec \ref{contrastive_loss}, ablation studies in Sec \ref{ablation}, and qualitative results in Sec \ref{QualitativeResults}. 

\begin{table*} 
    \centering 
  \caption{Automatic evaluation metrics on simple and hybrid text style transfer tasks. The reported models include our BTTS model with two different pretrained language model sizes (T5-base and T5-large), and previous work.} 
    \begin{subtable}{\linewidth}
    \centering
    \label{subtab:simple} 
    \begin{tabular}{c c c c c c c c} 
      \hline 
       & \multirow{2}{4em}{Method} & \multicolumn{3}{c}{Formal} & \multicolumn{3}{c}{Sentiment} \\
        \cmidrule(lr){3-5} \cmidrule(lr){6-8}
      & & \textbf{Accuracy} & \textbf{Content} & \textbf{G} & \textbf{Accuracy} & \textbf{Content} & \textbf{G} \\ 
      \hline 
      \multirow{4}{5em}{\textbf{Few-Shot}} & TextSETTR (T5-base) & 57.8 & 41.1 & 48.7 & 64.6 & 41.6 & 51.8 \\
      & TextSETTR (T5-large) & 62.5 & 43.5 & 52.1 & 67.1 & 44.8 & 54.8 \\
      & CP-G & 52.9 & 33.9 & 53.1 & 54.7 & 33.4 & 42.7 \\
      & CP-B & 35.3 & 39.5 & 35.8 & 51.8 & 29.5 & 39.1 \\
      \hline
      \multirow{2}{5em}{\textbf{Supervised}} & GST & 60.4 & 53.4 & 56.8 & 67.0 & 51.2 & 58.6 \\
      & DeleteAndRetrieve & 49.6 & 54.1 & 51.8 & 59.9 & 48.0 & 53.6 \\
      \hline 
      \multirow{2}{5em}{\textbf{Ours}} & BTTS (T5-base) & 60.1 & 52.8 & 56.3 & 64.9 & 51.5 & 57.8\\ 
      & BTTS (T5-large) & 66.3 & 52.9 & \textbf{59.2} & 70.4 & 53.5 & \textbf{61.4}  \\
      \hline
    \end{tabular}
    \hfill 
    \caption{Text style transfer tasks on simple datasets (formality and sentiment).}
    \end{subtable}

    \begin{subtable}{\linewidth}
    \centering
    \begin{tabular}{c c c c c c c c} 
      \hline 
       & \multirow{2}{4em}{Method} & \multicolumn{3}{c}{Genshin} & \multicolumn{3}{c}{Rephrase} \\
        \cmidrule(lr){3-5} \cmidrule(lr){6-8}
      & & \textbf{Accuracy} & \textbf{Content} & \textbf{G} & \textbf{Accuracy} & \textbf{Content} & \textbf{G} \\ 
      \hline 
      \multirow{4}{5em}{\textbf{Few-Shot}} & TextSETTR (T5-base) & 43.4 & 26.1 & 33.7 & 41.4 & 24.1 & 31.6 \\
      & TextSETTR (T5-large) & 47.7 & 29.3 & 37.4 & 45.7 & 27.3 & 35.3 \\
      & CP-G & 24.6 & 29.0 & 26.7 & 17.7 & 25.6 & 21.3    \\
      & CP-B & 20.5 & 36.9 & 27.5 & 19.5 & 32.1 & 25.0   \\
      \hline
      \multirow{2}{5em}{\textbf{Supervised}} & GST & 40.1 & 34.1 & 37.0 & 39.8 & 36.4 & 38.1  \\
      & DeleteAndRetrieve & 37.3 & 30.2 & 33.6 & 35.4 & 30.8 & 33.0  \\
      \hline 
      \multirow{2}{5em}{\textbf{Ours}} & BTTS (T5-base) & 45.5 & 38.2 & 41.7 & 43.5 & 36.2 & 39.7 \\ 
      & BTTS (T5-large) & 50.2 & 37.3 & \textbf{43.3} & 48.2 & 35.3 & \textbf{41.2}\\
      \hline
    \end{tabular}
    \hfill 
    \caption{Text style transfer tasks on complex datasets (Genshin and Rephrase).}
    \label{subtab:complex} 
    \end{subtable}
  \label{tab:main} 
\end{table*}
\subsection{Main Quantitative Results} 
\label{quantresult}
Table \ref{tab:main} compares the performances of our model against other state-of-art models. Our BTTS achieves the best performances in both classification accuracy and content preservation metrics among few shot models including CP-G and CP-B models by \cite{pmlr-v119-xu20a} and TextSETTR by \cite{riley2021textsettr}. Our model also exceeds by a small margin the performance of the models that utilize labeled data, including B-GST \cite{DBLP:journals/corr/abs-1908-09368}, DeleteAndRetrieve \cite{li-etal-2018-delete}, and CrossAligned \cite{https://doi.org/10.48550/arxiv.1705.09655}.

Table \ref{subtab:complex} shows the models' performances on \textit{Genshin} and \textit{Rephrase}. Both metrics significant dropped across all tested models. This is potentially due to the complex nature of the hybrid text style transfer on the two datasets. Our model again achieves the best performances among all the state-of-art models.

\subsection{Analysis on Contrastive Loss} 
We built BTTS upon the architecture of TextSETTR \cite{riley2021textsettr} with careful attention paid to ensure that the model architecture and parameters remained the same except for the addition of the contrastive loss module. As such, the experiment results presented in \ref{quantresult} naturally suffice for a comparative study on the effectiveness of the contrastive loss module. In this section, we provide further analysis of the contrastive loss module to provide additional insights into the workings of the contrastive loss and its potential for improving text style transfer performance.
\label{contrastive_loss}
\begin{figure*}
  \centering
  \begin{minipage}{0.15\linewidth}
    \centering
    \textbf{TextSETTR}\
  \end{minipage}%
  \begin{minipage}{0.35\linewidth}
    \centering
    \includegraphics[width=\linewidth]{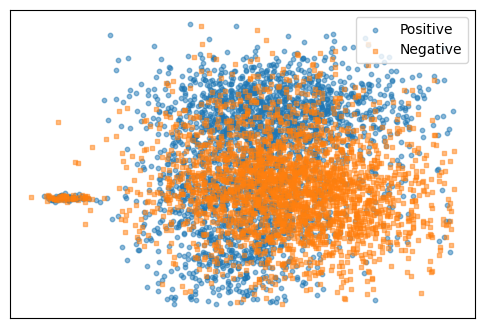}
  \end{minipage}%
  \begin{minipage}{0.35\linewidth}
    \centering
    \includegraphics[width=\linewidth]{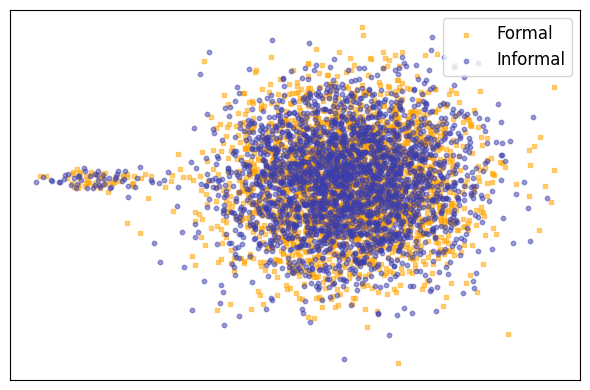}
  \end{minipage}
  \\ 
  \begin{minipage}{0.15\linewidth}
    \centering
    \textbf{BTTS}\
  \end{minipage}%
  \begin{minipage}{0.35\linewidth}
    \centering
    \includegraphics[width=\linewidth]{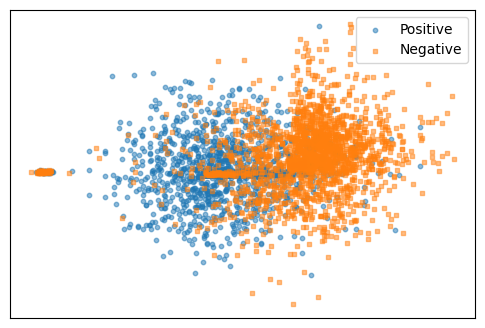}
  \end{minipage}%
  \begin{minipage}{0.35\linewidth}
    \centering
    \includegraphics[width=\linewidth]{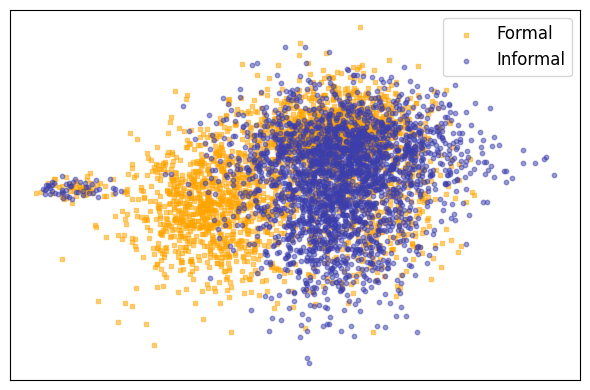}
  \end{minipage}
  \caption{2D UMAP embeddings of the style vectors learned by TextSETTR model (Top) and BTTS model (Bottom) on two datasets: Amazon review sentiment (Left) and Grammarly’s Yahoo Answers Formality Corpus (Right). BTTS model shows a better clustering and separation of the style vectors than TextSETTR model. However, perfect separation should not be expected as the dimensions have been compressed.}
\end{figure*}

\subsubsection{Hidden Embedding Visualization}
In order to show that our style extractor is capable of encoding various elements of textual style, we generated style vectors for 15,000 lines of text from three different review categories sourced from the Amazon data of \cite{ni-etal-2019-justifying}. We selected 2,500 positive (4 or 5 star) and 2,500 negative (1 or 2 star) samples, while removing examples where our BERT classifier disagreed with the label. In addition, we also selected 2,500 formal and 2,500 informal samples from the GYAFC dataset \cite{gyafcrao2018} to evaluate the ability of our style extractor in another perspective. The resulting 2D UMAP dimensionality reduction plot (Figure 3, bottom) clearly displays distinctions between sentiments. To compare, we also ran UMAP on style vectors from TextSETTR(Figure 3, top). The noticeable contrast between the two plots suggests that our training process helps to produce a representation space that distinguishes between the different attributes. 

\begin{figure}[t] 
    \centering 
    \includegraphics[width=0.8\linewidth]{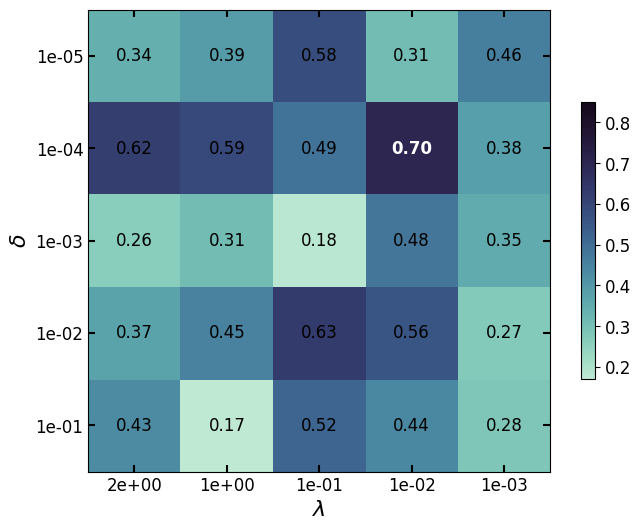} 
    \caption{Heatmap of BTTS's performance on the Amazon sentiment dataset with varying hyperparameter values in the Barlow Twins loss.} 
    \label{fig:param_sen} 
\end{figure}

\subsubsection{Parameter Sensitivity Analysis}

In this section, we perform a sensitivity analysis of our model's performance on the Amazon sentiment dataset with respect to two hyperparameters in the loss function, namely $\lambda$ and $\delta$ terms. The heatmap, displayed in Fig. \ref{fig:param_sen}, shows the model's accuracy on the Amazon sentiment dataset, as measured by ChatGPT, for 25 combinations of the two hyperparameters. The $\lambda$ term controls the relative weight between the Barlow Twins loss and the cross-entropy loss for the overall training objective in Eq. \ref{trainobj}, while the $\delta$ term controls the relative importance of the off-diagonal loss compared to the diagonal loss in Eq. \ref{barlowtwinsloss}. The heatmap demonstrates that the model's performance is highly sensitive to both hyperparameters, and that certain combinations of the two can lead to a significant improvement in accuracy, with the best setting at $\lambda = 1 * 10^{-2}$, $\delta = 1 * 10^{-4}$. 

\subsection{Ablation Studies}
\label{ablation}
\textbf{Model Size} In experiments, we study the impact of the language model size on the performance of BTTS by training the model with both T5-base and T5-large. T5-base consists of 220 million parameters while T5-large consists of 770 million parameters, and the results are shown in Table \ref{tab:main}. It shows that using a larger pretrained language model significantly increased the performance of our model, achieving state-of-art performances in all the style transfer tasks conducted in the experiment. 

\textbf{Exemplar Size} Our model sets up the inference procedure in a few-shot setting. In the main experiments, we use 30 sentences for each input and target text styles. In this section, we scrutinize the effect of the size of the exemplars on the performance of BTTS. We changed the size of the exemplars to 16, 8, 4, 2, 1, and 0 (representing zero-shot), and re-evaluated the model on Sentiment and Genshin datasets, whose results are reported in Table \ref{shotsizetable}. 
\begin{table}
\centering
\caption{Experiments on shot size. The model is based on BTTS(T5-large) and the $\lambda$ value is set to 4. The experiments are conducted during the inference stage, and hence no model training is involved.}
\begin{tabular}{ c c c c c } 
 \hline
  & \textbf{Shot Size} & \textbf{Acc.} & \textbf{Content} & \textbf{G} \\ 
 \hline
 {\textbf{Baseline}} & 30 &  70.4 & 53.5 & 61.4\\ 
 \hline
 \multirow{5}{2em}{\textbf{Few-Shot}}
 & 16 & 69.6 & 52.7 & 60.6\\
 & 8 & 69.3 & 53.1 & 60.7 \\
 & 4 & 68.0 & 49.6 & 58.1\\
 & 2 & 61.8 & 51.8 & 56.6\\
 & 1 & 56.5 & 48.3 & 52.2\\
 \hline
 \textbf{Zero-shot} & 0 & 54.2 & 51.3 & 52.7\\
 \hline
\end{tabular}
\label{shotsizetable}
\end{table}

The results show that the accuracy metric decreases somewhat insignificantly until the shot size becomes less than two. The experiments reveal BTTS' limitation on inference application under zero-shot or extremely few shot settings. However, it does tell that the model is still competitive with a fairly low amount of shot numbers. 

\subsection{Qualitative Results} 
\label{QualitativeResults}
\subsubsection{Human Evaluation}
\begin{table}
\centering
\begin{tabular}{ c c c c } 
 \hline
 \textbf{Method} & \textbf{Style} & \textbf{Content} & \textbf{Fluency} \\ 
 \hline
 BTTS (T5-base) &  2.3 & 2.9 & 3.2\\ 
 BTTS (T5-large) & 1.1 & 1.6 & 1.5\\
 \hline
 TextSETTR & 2.7 & 2.6 & 2.8 \\
 GST & 3.9 & 3.1 & 2.5\\
 \hline
\end{tabular}
\caption{Human Evaluation Metrics}
\label{HETable}
\end{table}
We conduct human evaluation as a complement to automatic metrics. We sample 50 examples for each of the four datasets (Sentiment, Formal, Genshin, Rephrase) and ask 30 experiment participants to rank the (1) style transferred strength (2) semantic preservation (3) sentence fluency of the four models. Model with the best performances is given the rank 1, whereas that of the worst performance is given the rank 4. 

Table \ref{HETable} shows the average ranking of the four models based on the responses from the experiment participants. It supports our claim that BTTS has the best performance on both single and hybrid style transfer tasks.

\subsubsection{Results Analysis}
\begin{table*}[t!]
    \caption{Examples of transferred sentences by BTTS and TextSETTR. Attributes are colored for formality and sentiment transfers on both directions.}
    \centering
    \begin{tabular}{p{2cm}|p{7cm}|p{7cm}}
    \hline
    & \textbf{{\color{red}Formal} $\Longrightarrow$ {\color{blue}Informal}} & \textbf{{\color{blue}Informal} $\Longrightarrow$ {\color{red}Formal}} \\
    \hline
    Input & 
    I {\color{red}do not} understand why people like films {\color{red}of that sort}. & 
    I think {\color{blue}pretty much} they're all {\color{blue}kind of} humorous. \\
    \hline
    TextSETTR & 
    I {\color{blue}no understand} why people like movie {\color{blue}like that}. & 
    I'm thinking they all funny, kind of. \\
    \hline
    BTTS & I {\color{blue}don't get} why people like those kinds of movies. & 
    I {\color{red}believe} they are all quite humorous.\\
    \hline
    \hline
    & \textbf{{\color{red}Positive} $\Longrightarrow$ {\color{blue}Negative}} & \textbf{{\color{blue}Negative} $\Longrightarrow$ {\color{red}Positive}} \\
    \hline
    Input 
    & I had it a long time now and I still {\color{red}love} it. 
    & I will {\color{blue}never} buy this product again. \\
    \hline
    TextSETTR 
    & I had it a {\color{blue}short} time, and I {\color{blue}never} loved it.
    & I not buy product again, but different product.\\
    \hline
    BTTS 
    & I had it for a long time now but I {\color{blue}no longer} love it. 
    & I will buy this product again.\\
    \hline
    \hline
    & \textbf{{\color{red}Hu Tao} $\Longrightarrow$ {\color{blue}Noelle}} & \textbf{{\color{blue}Noelle} $\Longrightarrow$ {\color{red}Hu Tao}} \\
    \hline
    Input & Lemme show you some fire tricks. First... Fire! And then... Whoosh! Fire butterfly! Be free! & I cannot rest on my laurels. After all, I'm not even formally a knight yet. Until that day, I must work even harder.\\
    \hline
    TextSETTR 
    & I'm gonna show you fire tricks. Fire! Whoosh! Butterfly! Free! & Can't relax on my laurels yet, need to work harder. After all, I am not even a knight yet.\\
    \hline
    BTTS 
    & Allow me to demonstrate some techniques with flames. First, observe as I ignite the fire into the form of a butterfly. It is now free to move. 
    & Oh, I simply cannot rest on my laurels! One day, I will officially become a knight. Until that day comes, I will dedicate myself to mastering my skills. I cannot wait to show the world just how magical I truly am! \\
    \hline
    \hline
    & \textbf{{\color{red}Simple} $\Longrightarrow$ {\color{blue}Creative}} & \textbf{{\color{blue}Creative} $\Longrightarrow$ {\color{red}Simple}} \\
    \hline
    Input & The American antelope, also called the pronghorn, is a typical grassland animal on the continent. & As a region grows in population, more infrastructure, such as roads, garbage dumps, and water treatment plants, will be required to support its residents.\\
    \hline
    TextSETTR 
    & The antelope from America, also called pronghorn, live on grassland like others.
    & When there's more people in a region, it means it needs more infrastructure, like roads, water treatment plants, and garbage dumps to keep everyone happy. \\
    \hline
    BTTS 
    & With its sleek frame and nimble hooves, the American antelope - also known as the pronghorn - is a quintessential denizen of the vast grasslands that sprawl across the continent. 
    & Growing population would mean more infrastructure like roads are required in a region. \\
    \hline
    \end{tabular}
    \label{tab:casestudy}
\end{table*}

Table \ref{tab:casestudy} shows a few examples of the transferred sentences on the four datasets in both transfer directions. We presents the inputs for each transfer case and the respective results of BTTS trained with T5-base and T5-large. We also include the outputs of TextSETTR for comparison. For \textit{Genshin}, we consider the style transfer between two specific characters, Hu Tao and Noelle. Hu Tao is regarded as a character with more active and outgoing personality, while Noelle is considered to be more calm and conservative. For \textit{Rephrase}, we consider the transfer task between simplicity and creativity. The same model is used for each case without extra training. 

\section{Related Works}
Most of the text style transfer methods take the line of unsupervised learning, where models are trained on text that have labeled attributes but are non-parallel. Various architecture have been utilized to learn the text style representation, including RNN \cite{li-etal-2018-delete} and Transformers \cite{DBLP:journals/corr/abs-1908-09368}. While these methods have shown great success, their applicability are limited by the style labels that are required for training, which are not readily available for many desired attributes in the real world.  

Recently, \cite{pmlr-v119-xu20a} develops a few shot approach that trains on unlabeled data, and only takes a few samples during inference for style transfer. This line of work removes the need for labeled training samples, but generally do not perform well on complicated style transfer tasks involving transferring multiple attributes simultaneously. Our work is most closely related to \cite{riley2021textsettr}, which includes an extractor module consisting of a large language model to learn style representations. Our work differs in the way that we add Barlow Twins loss to improve the robustness of the model in complex style transfer tasks, and also achieve state-of-art performances on one-to-one attribute transfer. 

The current advancement in large language models (LLMs) have shown remarkable capabilities in generating natural language texts across various domains and tasks. Some works have explored the use of LLMs for zero-shot text style transfer, where no model fine-tuning or exemplars in the target style are required. Instead, a natural language instruction is given to the LLM as a prompt, and the LLM is expected to rewrite the input sentence in the desired style.
One such work is \cite{reif-etal-2022-recipe}, who propose a augmented zero-shot learning method for arbitrary text style transfer with LLMs. They frame style transfer as a sentence rewriting task and use a natural language instruction that specifies both the input and output styles, as well as some additional information such as synonyms or antonyms. They show that their method can perform well on standard style transfer tasks such as sentiment and formality, as well as on more creative transformations such as “make this melodramatic” or “insert a metaphor”. They also demonstrate that their method can handle multiple styles simultaneously and generate diverse outputs.

\section{Conclusion}
We formally defined the concept of complex text style transfer and constructed a large-scale dataset for the task. We have explored the use of small models with implicit style pre-training, which achieved state-of-art performances of few-shot approaches on complex text style transfer tasks. We have introduced a novel evaluation metric based on ChatGPT that has better alignment with human judgement.

\clearpage
\bibliography{ecai}
\end{document}